\documentclass[11pt]{article}

\usepackage[preprint]{acl}

\usepackage{times}
\usepackage{latexsym}

\usepackage[T1]{fontenc}

\usepackage[utf8]{inputenc}

\usepackage{microtype}

\usepackage{inconsolata}

\usepackage{graphicx}
\usepackage{verbatim} 
\usepackage{enumitem} 
\usepackage{amsmath}
\usepackage{amssymb}  
\usepackage{booktabs}
\usepackage{tcolorbox}
\usepackage{tabularx}
\usepackage{booktabs} 
\usepackage{booktabs}  
\usepackage{multirow}  
\usepackage{booktabs}
\usepackage{amssymb}
\usepackage[table]{xcolor}
\usepackage{tabularx}

\definecolor{elderblue}{RGB}{245,250,255}

\definecolor{gain}{RGB}{26,127,74}
\definecolor{gainbg}{RGB}{235,247,239}

\newcommand{\gain}[1]{%
  \cellcolor{gainbg}\textcolor{gain}{\bfseries +#1}%
}

\newcommand{\cmark}{\ensuremath{\checkmark}}
\newcommand{\xmark}{\ensuremath{\times}}
\newcommand{\pmark}{\ensuremath{\triangle}}

\title{ElderBench: Benchmarking Autonomous Mobile Agents for Older Adults}

\author{%
\normalfont\mdseries\small
\begin{tabular*}{0.96\textwidth}{@{\extracolsep{\fill}}ccc@{}}
\begin{tabular}{@{}c@{}}
{\large Weide Zhan} \\[0.25em]
Fudan University \\
Shanghai, China \\
wdzhan25@m.fudan.edu.cn
\end{tabular}
&
\begin{tabular}{@{}c@{}}
{\large Qumu Shaqu} \\[0.25em]
Fudan University \\
Shanghai, China \\
25213050317@m.fudan.edu.cn
\end{tabular}
&
\begin{tabular}{@{}c@{}}
{\large Yuanqing Liu} \\[0.25em]
Fudan University \\
Shanghai, China \\
25213050283@m.fudan.edu.cn
\end{tabular}
\\[2.5em]
\begin{tabular}{@{}c@{}}
{\large Peng Zhang} \\[0.25em]
Fudan University \\
Shanghai, China \\
zhangpeng\_@fudan.edu.cn
\end{tabular}
&
\begin{tabular}{@{}c@{}}
{\large Jiahao Liu} \\[0.25em]
Fudan University \\
Shanghai, China \\
jiahaoliu21@m.fudan.edu.cn
\end{tabular}
&
\begin{tabular}{@{}c@{}}
{\large Kam Him Lam} \\[0.25em]
Fudan University \\
Shanghai, China \\
24302010074@m.fudan.edu.cn
\end{tabular}
\\[2.5em]
\begin{tabular}{@{}c@{}}
{\large Ning Gu} \\[0.25em]
Fudan University \\
Shanghai, China \\
ninggu@fudan.edu.cn
\end{tabular}
&
\begin{tabular}{@{}c@{}}
{\large Zhan Hu} \\[0.25em]
Fudan University \\
Shanghai, China \\
huzhan@fudan.edu.cn
\end{tabular}
&
\begin{tabular}{@{}c@{}}
{\large Tun Lu} \\[0.25em]
Fudan University \\
Shanghai, China \\
lutun@fudan.edu.cn
\end{tabular}
\end{tabular*}%
}

\begin{document}

\maketitle
\begin{abstract}
While autonomous mobile agents hold great potential for assisting older adults with smartphone usage, existing GUI benchmarks mainly rely on explicit, goal-oriented instructions and rarely capture the naturally occurring language patterns of older users, such as indirect speech, referential ambiguity, and under-specified requests. This mismatch between benchmark instructions and real-world elderly interactions may hinder reliable agent deployment. To address this gap, we present ElderBench, the first benchmark for evaluating mobile GUI agents in authentic elderly-oriented scenarios. ElderBench is constructed from 249 naturally elicited smartphone tasks collected from older adults across 20 applications. We first characterize the linguistic divergence between elderly instructions and existing GUI benchmark instructions from syntactic, semantic, and pragmatic perspectives. We then evaluate mainstream GUI agents and Vision-Language Models under both online and offline settings, revealing substantial performance degradation when handling elderly-oriented instructions. Through controlled instruction normalization, failure analysis, and fine-grained linguistic feature analysis, we further identify how elderly-specific language patterns contribute to agent failures. Our findings provide actionable design insights toward more adaptive, interpretable, and age-inclusive GUI agents for older adults.

\end{abstract}

\section{Introduction}
Recent advances in large language models (LLMs) have substantially expanded the capabilities of intelligent assistants, particularly in language understanding, reasoning, planning, and tool use \cite{yao2022react,shu2024rah}. Building on these capabilities, LLM-based agents are evolving beyond conventional conversational systems toward autonomous assistants that can maintain contextual information, reason over user needs, and interact with external tools and digital services \cite{liu2025agentcf++,liu2025filtering}. Recent studies further suggest that such assistants may increasingly serve as intermediaries between users and everyday digital services, supporting activities such as search, shopping, travel planning, scheduling, recommendation, and content access \cite{liu2026hidden}. This shift moves human--computer interaction beyond merely responding to user queries toward interpreting user intent and translating it into concrete actions in digital environments.

Graphical User Interface (GUI) agents represent an important realization of this emerging paradigm. By combining multimodal perception, reasoning, and action generation, GUI agents can directly operate digital interfaces through human-like actions such as clicking, swiping, and typing, allowing users to accomplish smartphone tasks through natural-language instructions \cite{zhang2025appagent,wang2023enabling,hong2024cogagent,rawles2023androidinthewild}. Recent advances in Vision-Language Models (VLMs) have further strengthened agents' ability to perceive interface states, identify actionable elements, and plan multi-step interaction trajectories \cite{openai2023gpt,anthropic2023introducing,glm2024chatglm}. Compared with traditional conversational assistants that primarily provide textual guidance, GUI agents can directly carry out interface operations on behalf of users, potentially reducing the need to understand application-specific procedures.

This capability is particularly valuable for older adults, who often encounter difficulties in smartphone usage due to limited ICT literacy, unfamiliar interface conventions, and challenges in understanding complex digital procedures \cite{korpela2023older,leung2012older,pang2015older}. Tasks such as online shopping, transportation booking, and mobile payment, which may be routine for younger users, can therefore become substantial barriers for older adults. By allowing users to express their needs in natural language and delegating the corresponding interface operations to an agent, GUI agents offer a promising approach to lowering these barriers while preserving a visible and inspectable interaction process.

However, whether current GUI agents can reliably assist older adults remains unclear. Existing GUI agent benchmarks, such as AndroidWorld \cite{rawles2025androidworld}, MobileWorld \cite{kong2026mobileworld}, and AndroidLab \cite{xu2025androidlab}, primarily evaluate agents using explicit, goal-oriented, and carefully specified instructions. For example, a typical benchmark instruction may specify the exact target, location, and required operation: \textit{"Find the resume file downloaded within the last month in the Download folder, and send it to HR\_chen@gmail.com with the subject candidates\_cv."} Although such instructions facilitate reliable evaluation, they do not reflect how older adults naturally express smartphone needs. In real-world interactions, older users often provide indirect requests, incomplete descriptions, referential expressions, and non-standard language patterns due to limited technical vocabulary and unfamiliarity
with digital interfaces \cite{sharifi2025older}. This discrepancy between benchmark instructions and elderly users' natural expressions creates a critical evaluation gap for age-inclusive GUI agents.

Despite the importance of this problem, existing benchmarks provide limited support for evaluating GUI agents in elderly-oriented scenarios, mainly due to the lack of naturally collected older-adult interaction data. This motivates three research questions: \textbf{RQ1:} How do naturally elicited elderly GUI instructions differ from conventional GUI benchmark instructions? \textbf{RQ2:} How well can current GUI agents and VLMs execute tasks described through elderly-oriented instructions? \textbf{RQ3:} What linguistic and execution-level factors contribute to agent failures in elderly scenarios?

To answer these questions, we introduce \textbf{ElderBench}, the first benchmark designed for evaluating mobile GUI agents under authentic elderly-oriented instructions. We collect 249 naturally elicited smartphone tasks from older adults through semi-structured interviews, covering 20 Android applications. ElderBench provides (1) linguistic characterization of elderly GUI instructions, (2) online and offline evaluation protocols for measuring agent execution performance, and (3) analysis tools including controlled instruction normalization, failure analysis, and linguistic factor analysis to understand the causes of agent failures.

Our contributions are summarized as follows:
\begin{itemize}
    \item We introduce the first real-world elderly-oriented GUI instruction benchmark, consisting of 249 naturally elicited smartphone tasks collected from older adults. We systematically characterize their linguistic differences from conventional GUI instructions across syntactic, semantic, and pragmatic dimensions.
    
    \item We establish ElderBench, an evaluation framework for mobile GUI agents and VLMs under elderly-oriented scenarios. Through online and offline evaluations, we reveal substantial limitations of current agents in handling naturally occurring older-adult instructions.

    \item We conduct controlled normalization, failure analysis, and linguistic factor analysis to identify the sources of agent failures and derive practical design insights for developing more adaptive and age-inclusive GUI agents.
\end{itemize}

\begin{table}[t]
\centering
\footnotesize
\setlength{\tabcolsep}{1.5pt}

\caption{Comparison with representative mobile GUI-agent benchmarks. TU denotes target-user elicitation; PS denotes population-specific design; Nat. denotes naturally elicited user language phenomena; and LA denotes systematic linguistic analysis. \cmark, \pmark, and \xmark indicate full, partial, and no support, respectively.}
\label{tab:benchmark_comparison}

\resizebox{\columnwidth}{!}{
\begin{tabular}{@{}lrrcccc@{}}
\toprule
\textbf{Benchmark}
& \textbf{Apps}
& \textbf{Tasks}
& \textbf{TU}
& \textbf{PS}
& \textbf{Nat.}
& \textbf{LA} \\
\midrule

AndroidWorld \cite{rawles2025androidworld}
& 20 & 116
& \xmark & \xmark & \xmark & \xmark \\

AndroidLab \cite{xu2025androidlab}
& 9 & 138
& \xmark & \xmark & \xmark & \xmark \\

A3 \cite{chai2026a3}
& 20 & 100
& \xmark & \xmark & \xmark & \xmark \\

ProBench \cite{yang2026probench}
& 34 & 217
& \xmark & \xmark & \xmark & \xmark \\

MobileWorld \cite{kong2026mobileworld}
& 20 & 201
& \xmark & \xmark & \pmark & \xmark \\

KnowU-Bench \cite{chen2026knowu}
& 23 & 192
& \xmark & \cmark & \pmark & \xmark \\

\midrule
\rowcolor{elderblue}
\textbf{ElderBench}
& \textbf{20}
& \textbf{249}
& \cmark & \cmark & \cmark & \cmark \\

\bottomrule
\end{tabular}
}

\end{table}


\section{Related Work}
\subsection{GUI Agents}
Driven by the rapid advancement of Large Foundation Models (LFMs), AI systems have evolved from conventional conversational assistants toward autonomous agents capable of planning and executing complex tasks \cite{chen2024survey,cheng2024exploring,zhao2023depth,yao2022react}.
Among these emerging paradigms, GUI agents have attracted increasing attention due to their ability to directly interact with digital environments through human-like actions, such as tapping, swiping, and typing, across smartphones, desktops, and web platforms \cite{lu2024omniparser,iong2024openwebagent,hong2024cogagent,nguyen2025gui}. 

GUI agents provide a promising approach for assisting older adults by executing operations directly on interfaces rather than requiring users to understand complicated procedures. Their visible interaction process also improves transparency and preserves users' sense of control when delegating smartphone tasks \cite{zhang2025wepilot}.

The effectiveness of GUI agents largely depends on the visual understanding and reasoning capabilities of Vision-Language Models (VLMs). OmniParser \cite{lu2024omniparser} explores VLM-based UI parsing by transforming screenshots into structured interface elements. AutoGLM \cite{liu2024autoglm} further incorporates multimodal context understanding and strategic planning to automate user tasks. UI-TARS \cite{wang2025ui} investigates large-scale VLM-based GUI interaction through reinforcement learning. Specifically targeting older adults, WePilot \cite{zhang2025wepilot} leverages visual understanding capabilities to assist older users in completing smartphone operations and reducing digital barriers.

\subsection{GUI Agent Benchmarks}
Reliable benchmarks are essential for evaluating the capabilities and limitations of GUI agents. Existing benchmarks mainly follow two evaluation paradigms: online interaction-based evaluation and offline trajectory-based evaluation.

Online benchmarks evaluate agents in executable environments where actions dynamically change system states. In mobile scenarios, AndroidWorld \cite{rawles2025androidworld} provides a functional Android environment with programmatically generated tasks across real-world applications, AndroidLab \cite{xu2025androidlab} introduces an infrastructure for evaluating multimodal agents on mobile devices, and MobileWorld \cite{kong2026mobileworld} further explores long-horizon cross-application interaction in a virtual emulator environment.

Offline benchmarks instead evaluate agents using pre-recorded trajectories and annotated GUI states. AitW \cite{rawles2023androidinthewild} provides large-scale human demonstrations containing screenshots, actions, and task goals, while AMEX \cite{chai2025amex} enriches offline evaluation with multi-level GUI annotations.

Beyond standard task completion, recent benchmarks have begun exploring more realistic interaction challenges, including incomplete instructions and ambiguous user intents. Mobile-Bench-v2 \cite{xu2025mobile} introduces an ambiguous instruction setting by removing task-specific slots from complete instructions and evaluating whether agents can recover missing information or initiate clarification. These studies demonstrate the importance of moving beyond fully specified instructions toward uncertain user-agent interactions.

As summarized in Table~\ref{tab:benchmark_comparison}, existing mobile GUI benchmarks provide broad application and task coverage, but rarely combine target-user elicitation, population-specific design, naturally occurring language phenomena, and systematic linguistic analysis. ElderBench contains 249 tasks across 20 applications, representing the largest task collection among the compared benchmarks. More importantly, its instructions are directly elicited from older adults rather than constructed or modified from predefined task templates, enabling the evaluation of indirect speech, referential ambiguity, disfluency, and under-specification in authentic usage scenarios.

\subsection{Language Characteristics of Older Adults}

Older adults increasingly rely on digital technologies for communication, healthcare, transportation, and daily services. However, many older adults still use only a limited subset of available applications and functions due to restricted digital literacy and insufficient technical support \cite{gelderblom2010mobile,li2016older,li2018understanding,olphert2005towards,barnard2013learning,kim2020comparison,pang2021technology}.

Beyond operational difficulties, prior studies have shown that older adults exhibit distinctive communication patterns when describing technology-related needs. Compared with younger users, older adults are more likely to rely on contextual descriptions, fragmented expressions, and ambiguous references when seeking technical assistance \cite{sharifi2025older,cho2021lexical}. These language characteristics create additional challenges for autonomous agents that typically assume explicit and standardized user commands.

Although GUI agents provide a promising approach for reducing digital barriers among older adults, existing evaluation frameworks rarely consider how older users naturally formulate their intentions. ElderBench therefore complements previous GUI benchmarks by focusing on the linguistic challenges arising from authentic elderly-oriented interactions.


\section{Data Collection and Linguistic Analysis}
\subsection{Data Collection}
\textbf{Participant Recruitment.} We recruited a heterogeneous cohort of 28 older adults aged between 59 and 84 years old to capture diverse linguistic expressions and smartphone usage patterns among older users. The participants consisted of 8 males (28.57\%) and 20 females (71.43\%), covering different age ranges within the older adult population. Beyond demographic diversity, participants exhibited substantial variation in smartphone familiarity and usage habits: some participants had more than ten years of smartphone experience, while others had only 1--8 years of experience. Their daily smartphone engagement also varied considerably, ranging from light users with approximately 1--2 hours of daily usage to highly engaged users spending more than 5--6 hours per day on mobile devices.
To further increase contextual diversity, participants were recruited from multiple provinces and municipalities in China, including Shanghai, Anhui, Jiangsu, Hebei, and other regions. 
Before collecting task instructions, we explained the research objectives and anonymization policy for raw data. All data collection activities were conducted upon receiving verbal informed consent from participants. Demographics are summarized in supplements.

\textbf{Data Collection.} To capture naturally occurring elderly GUI instructions, we conducted semi-structured interviews \cite{pierres2025exploring,liu2025envisioning} without predefined task categories or linguistic examples. The interviews consisted of two phases. First, participants described their routine smartphone activities and digital tasks encountered in daily life. After introducing the concept of GUI agents, participants discussed how such agents could assist their existing smartphone usage. Second, we explored additional scenarios where participants experienced difficulties or desired assistance, allowing them to describe tasks they would like agents to perform.

After removing two invalid instructions through discussion among three authors, we obtained 249 valid elderly GUI instructions from 251 collected instructions.

\subsection{Data Analysis}

To characterize how elderly GUI instructions \textit{(ElderBench's instructions)} differ from conventional GUI instructions, we compare ElderBench instructions with MobileWorld instructions \cite{kong2026mobileworld} as a representative baseline. Following linguistic theory \cite{silverstein1972linguistic,kasirzadeh2023conversation}, we analyze instructions from three dimensions: syntax, semantics, and pragmatics. We report the main findings below, while complete statistics are provided in the supplementary material.

\subsubsection{Syntactic Dimension}
Grounded in linguistic theory \cite{silverstein1972linguistic,kasirzadeh2023conversation}, we analyze instruction length, predicate density (verb counts), and structural patterns, considering that GUI tasks are fundamentally action-oriented. Compared with baseline instructions, which are mainly explicit and action-driven descriptions designed for complex workflows, ElderBench instructions exhibit shorter and more heterogeneous structures.

Conversely, ElderBench's instructions are uniquely characterized by ``minimalist'' and multi-dimensional heterogeneity. Specifically, 78.31\% of ElderBench instructions contain fewer than 20 characters. Moreover, while baseline instructions are dominated by compound structures, elderly instructions are distributed across multiple syntactic patterns. Elliptical and narrative structures, which rarely appear in baseline instructions, account for 47.80\% of elderly instructions.

This structural gap indicates that elderly-oriented GUI scenarios introduce challenges beyond long-horizon planning: agents must first infer user intent from short, incomplete, and non-canonical expressions before generating executable actions.

\begin{figure*}[t]
    \centering
    \includegraphics[width=1\textwidth]{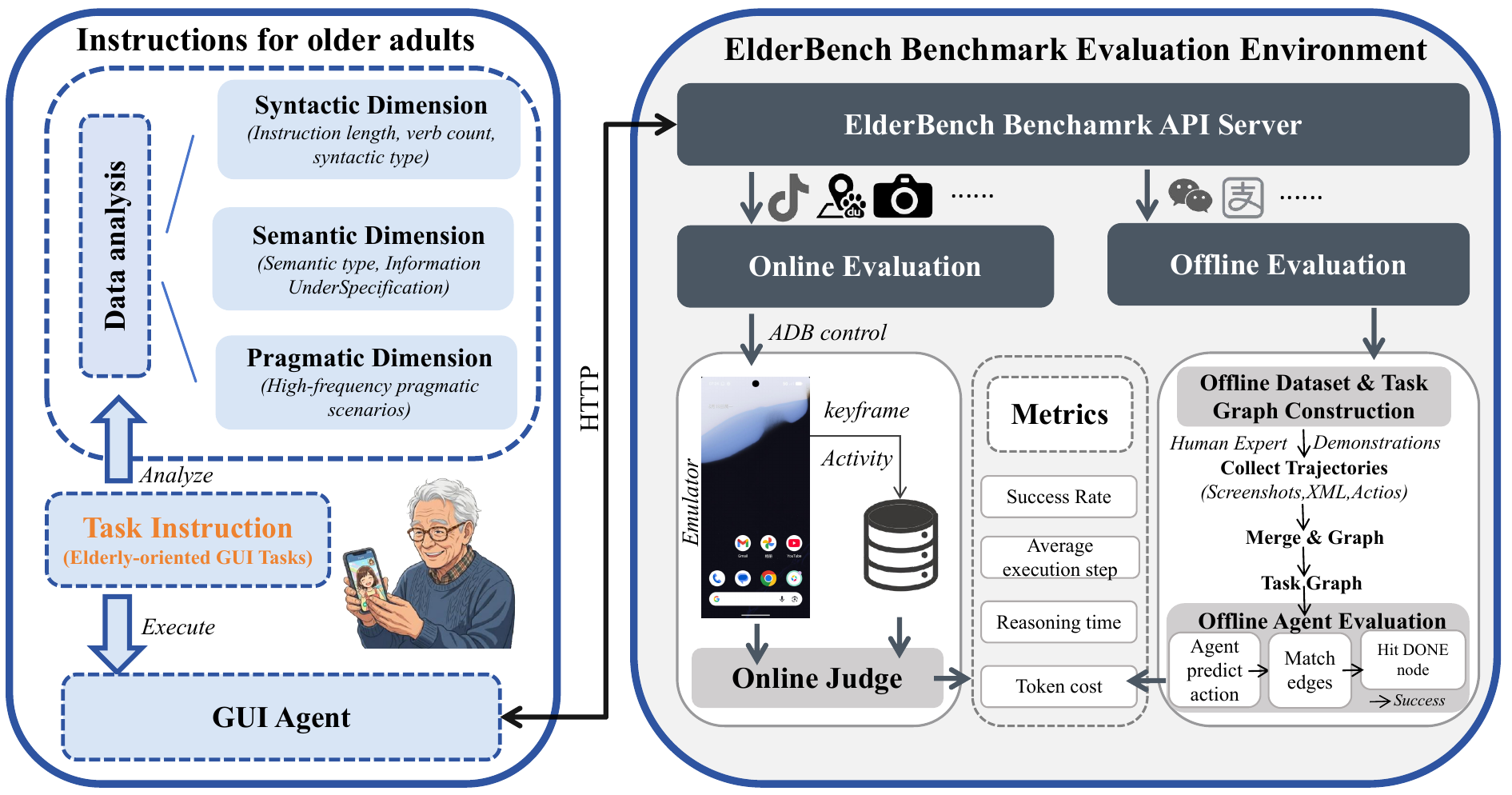}
    \caption{Overview of the ElderBench benchmarking framework. The architecture has a dual-module assessment mode supporting both Online and Offline evaluation modes to systematically evaluate autonomous mobile agents under elderly GUI instructions.}
    \label{fig:framework}
\end{figure*}

\subsubsection{Semantic Dimension}

We perform thematic analysis
\cite{clarke2017thematic} to identify semantic differences between ElderBench and baseline instructions. We categorize elderly instructions into five phenomena: Clear and Explicit, Disfluency/Correction, Indirect Speech, Referential Ambiguity, and Under-specification (definitions and complete distributions are provided in the supplementary material).

\textbf{Implicit state descriptions.} Baseline instructions primarily specify explicit actions, resulting in 0.00\% occurrence of Disfluency/Correction, Referential Ambiguity, and Indirect Speech. In contrast, ElderBench contains 19.28\% Indirect Speech examples, such as ``The sound is too low, I cannot hear it.'' Among these instructions, 22.92\% contain no explicit verbs, suggesting that older adults often describe desired states or difficulties rather than directly specifying operational actions.

\textbf{Context-dependent references.} ElderBench also contains 12.85\% Referential Ambiguity cases, where users rely on expressions anchored in the current interface context, such as ``Add this number to the blacklist'' or ``Screenshot this page.'' Such instructions cannot be reliably interpreted from text alone, highlighting the importance of visual grounding for GUI agents. GUI agents must be capable of associating ambiguous pronouns such as \textit{this} and \textit{that} with specific spatial UI components or text blocks rendered in real time on the active screen.

\subsubsection{Pragmatic Dimension}

From a pragmatic perspective, we analyze the application scenarios and functional demands reflected in elderly instructions. Elderly users' requests are mainly concentrated in Life Services (e.g., e-commerce, digital payment, and transportation), Social Communication, and System Settings. These scenarios correspond to essential daily activities and accessibility needs, such as font adjustment, storage management, and ringtone customization. \textit{(Complete scenario distributions appear in supplements.)}

\section{Benchmark}
To systematically evaluate GUI agents in elderly-oriented smartphone scenarios, we construct \textbf{ElderBench}, a benchmark consisting of 249 naturally collected tasks across 20 Android applications. ElderBench adopts a dual-mode evaluation framework (Figure~\ref{fig:framework}) to balance real-world interaction fidelity and evaluation controllability.

\subsection{Operating Environment}
ElderBench focuses on single-turn GUI task execution, where each instance contains one user instruction from an initial device state. This design is motivated by three factors: first, as the first elderly-oriented GUI benchmark, we aim to establish a fundamental evaluation setting for naturally occurring older-adult instructions; second, our interviews show that older adults frequently express digital needs as independent task requests, reflecting difficulties in maintaining complex interaction contexts under limited ICT literacy; Third, from a technical perspective, most current mobile GUI agents still operate primarily on single-turn user instructions; therefore, single-turn evaluation best reflects the present state of the technology.

Following previous GUI agent studies \cite{liu2024autoglm}, we define the action space as human-like operations, including Tap, Long Press, Swipe, Type, Home, and Back. All coordinates are normalized into the range [0,1000] and automatically mapped to device-specific pixel coordinates.

ElderBench provides two complementary evaluation modes: Online evaluation and Offline evaluation. The online setting measures closed-loop interaction in live environments, while the offline setting provides a reproducible evaluation protocol based on human-demonstrated task graphs. Together, they cover 249 tasks across 20 applications from elderly daily usage scenarios. 

\subsection{Evaluation Protocol}

ElderBench adopts two complementary evaluation protocols: online and offline evaluation. The protocol assignment is determined at the task level based on execution reproducibility, privacy constraints, and environment stability.
The online subset contains 130 tasks across 15 applications. These tasks are executed in a closed-loop environment, where agents directly interact with live applications and receive updated GUI states after each action. The offline subset contains 119 tasks across five applications, including payment, communication, ride-hailing, shopping, and travel scenarios. These tasks are evaluated against preconstructed task graphs derived from human demonstrations, avoiding privacy-sensitive operations and unstable external factors.

\subsubsection{Online Evaluation}

The online evaluation mode evaluates agents through direct interaction with Android environments. Agents perceive screenshots and contextual information, generate structured actions, and execute them through Android Debug Bridge (ADB), forming a closed-loop perception-action cycle.

A task is considered successful only when both agent termination and task verification are satisfied:
\textit{$\text{Success} = \text{Agent\_Finished} \cap \text{Judge\_Passed}$}.

For automated verification, we adopt a VLM-as-a-Judge paradigm inspired by \cite{shih2026judge}. Specifically, Qwen3-VL-Flash is used as a zero-shot evaluator based on three inputs: (1) the final screenshot, (2) sampled intermediate frames, and (3) the active Android Activity. To validate the reliability of automated evaluation, we manually verified AutoGLM results and obtained 93.08\% agreement between human judgment and the VLM judge.

\subsubsection{Offline Evaluation}

To improve reproducibility and avoid uncontrolled factors in partially real-world applications (e.g., account states, dynamic content, and authentication restrictions), we construct an offline evaluation set based on human demonstrations, following previous offline GUI benchmarks \cite{xu2025mobile,rawles2023androidinthewild,chai2025amex}.

For each offline task, human annotators execute the target instruction on physical devices and record the interaction trajectory: $(O_t,V_t,A_t)$, where $O_t$ represents the screenshot, $V_t$ represents the UI hierarchy (XML), and $A_t$ represents the executed action (e.g., Tap, Swipe, Type). Multiple valid trajectories are collected for the same task to avoid dependence on a single golden path.

The collected trajectories are merged into task graphs. Specifically, we extract UI signatures, anchor texts, package information, and component-level features to identify equivalent states. Matched states are merged as graph nodes, while human-executed actions are represented as directed edges. During evaluation, an agent succeeds if its predicted action sequence reaches the terminal node of the task graph. Click actions allow a 50-pixel tolerance, and text inputs support fuzzy matching. The task is only considered complete when the agent explicitly outputs \textit{finish(...)} and reaches the predefined DONE state.


\subsubsection{Evaluation metrics}
To provide a multi-dimensional assessment of mobile GUI agents under elderly-oriented scenarios, we establish an evaluation suite: Task Success Rate, Average Inference Latency and Monetary Cost Metric:
\paragraph{(1) Task Success Rate (SR):}
The ratio of successfully completed tasks to total tasks:
\begin{equation}
\text{SR} = \frac{1}{N} \sum_{i=1}^{N} \mathbb{I}(\text{Success}_i)
\end{equation}
where $N$ is the total task count, and indicator $\mathbb{I}(\cdot) = 1$ if task $i$ satisfies the online/offline completion criteria, and $0$ otherwise.

\paragraph{(2) Average Inference Latency (AIL):}
Measures the wall-clock duration per operational turn to capture temporal burden. 
For an evaluation subset $\mathcal{D}_m$, where $m \in \{\mathrm{online}, \mathrm{offline}, \mathrm{overall}\}$, it is defined as:
\begin{equation}
\mathrm{AIL}_m =
\frac{
\sum_{i \in \mathcal{D}_m}
\sum_{t=1}^{T_i}
\left(
\Delta \tau_{\mathrm{reasoning}}^{(i,t)}
+
\Delta \tau_{\mathrm{parsing}}^{(i,t)}
\right)
}{
\sum_{i \in \mathcal{D}_m} T_i
},
\end{equation}

where $T_i$ denotes the total operation steps for task $i$. $\Delta \tau_{\mathrm{reasoning}}^{(i,t)}$ and $\Delta \tau_{\mathrm{parsing}}^{(i,t)}$ denote the model inference time and structured action-parsing overhead at step $t$, respectively.

\paragraph{(3) Monetary Cost Metric (MCM):}
Quantifies deployment cost by tracking average token consumption per task:
\begin{equation}
\text{MCM} = \frac{1}{N} \sum_{i=1}^{N} \left(  \mathcal{K}_{\text{prompt}}^{(i)} + \mathcal{K}_{\text{gen}}^{(i)} \right)
\end{equation}
where $\mathcal{K}_{\text{prompt}}$ and $\mathcal{K}_{\text{gen}}$ are the quantities of consumed input and output tokens for task $i$.

\begin{table*}[t]
\centering
\small
\setlength{\tabcolsep}{2.6pt}
\caption{
Performance on ElderBench under the overall, online, and offline evaluation protocols. Online evaluation contains 130 tasks across 15 applications, while offline evaluation contains 119 tasks across 5 applications. SR denotes Task Success Rate (\%), AIL denotes Average Inference Latency per operational turn (seconds), and MCM denotes average token consumption per task. The best result within each model category and evaluation mode is highlighted in bold.
}
\label{tab:mode_breakdown}
\resizebox{\textwidth}{!}{
\begin{tabular}{ll
ccc
ccc
ccc}
\toprule
\multirow{2}{*}{\textbf{Architecture}} &
\multirow{2}{*}{\textbf{Model}} &
\multicolumn{3}{c}{\textbf{Overall (249)}} &
\multicolumn{3}{c}{\textbf{Online (130)}} &
\multicolumn{3}{c}{\textbf{Offline (119)}} \\
\cmidrule(lr){3-5}
\cmidrule(lr){6-8}
\cmidrule(lr){9-11}
& &
\textbf{SR$\uparrow$} &
\textbf{AIL$\downarrow$} &
\textbf{MCM$\downarrow$} &
\textbf{SR$\uparrow$} &
\textbf{AIL$\downarrow$} &
\textbf{MCM$\downarrow$} &
\textbf{SR$\uparrow$} &
\textbf{AIL$\downarrow$} &
\textbf{MCM$\downarrow$} \\
\midrule

\multirow{5}{*}{VLMs}
& Qwen3-VL-Flash
& 22.89 & \textbf{3.57} & 66,657
& 40.00 & \textbf{3.74} & 118,347
& 4.20 & \textbf{1.99} & \textbf{10,189} \\

& GLM-4.6V
& 23.29 & 13.08 & 25,768
& 33.85 & 13.11 & \textbf{30,465}
& 11.76 & 13.04 & 20,638 \\

& Kimi K2.5
& 29.72 & 7.40 & 52,900
& 43.85 & 6.82 & 87,732
& 14.29 & 10.45 & 14,849 \\

& Qwen3-VL-Plus
& 47.39 & 6.66 & 35,260
& \textbf{76.15} & 7.57 & 57,336
& 15.97 & 3.59 & 11,144 \\

& Gemini-3-Flash
& \textbf{49.80} & 8.00 & \textbf{21,660}
& 70.00 & 8.87 & 30,572
& \textbf{27.73} & 5.67 & 11,925 \\
\midrule

\multirow{4}{*}{\shortstack[l]{GUI\\Agents}}
& Doubao-1.5-UI-TARS (7B)
& 32.53 & 7.24 & \textbf{29,741}
& 53.08 & \textbf{6.78} & 48,239
& 10.08 & 9.68 & \textbf{9,534} \\

& GUI-Owl-1.5 (8B-Think)
& \textbf{45.38} & 8.00 & 66,293
& \textbf{68.46} & 9.77 & 107,507
& 20.17 & \textbf{2.64} & 21,271 \\

& UI-Venus-1.5 (8B)
& 41.37 & \textbf{6.44} & 35,847
& 50.77 & 7.55 & \textbf{45,764}
& \textbf{31.09} & 3.14 & 25,015 \\

& AutoGLM (Phone-9B)
& 36.55 & 6.97 & 43,680
& 51.54 & 7.05 & 65,150
& 20.17 & 6.70 & 20,224 \\
\bottomrule
\end{tabular}
}
\end{table*}

\begin{table}[t]
\centering
\small
\setlength{\tabcolsep}{1.8pt}
\caption{
Successful tasks before and after normalization on a stratified 100-task subset. ``Orig.'' and ``Rewr.'' denote the original and rewritten instructions, and $\Delta$ their difference.
}
\label{tab:instruction_normalization}
\resizebox{\columnwidth}{!}{
\begin{tabular}{l rr cc>{\columncolor{gainbg}}c
                         cc>{\columncolor{gainbg}}c}
\toprule
\multirow{2}{*}{\textbf{Category}} &
\multirow{2}{*}{\textbf{Share (\%)}} &
\multirow{2}{*}{\textbf{\#Tasks}} &
\multicolumn{3}{c}{\textbf{AutoGLM (Phone-9B)}} &
\multicolumn{3}{c}{\textbf{Qwen3-VL-Flash}} \\
\cmidrule(lr){4-6}
\cmidrule(lr){7-9}
& & &
\textbf{Orig.} &
\textbf{Rewr.} &
$\boldsymbol{\Delta}$ &
\textbf{Orig.} &
\textbf{Rewr.} &
$\boldsymbol{\Delta}$ \\
\midrule

Clear and Explicit
& 17.67 & 18
& 5 & 9 & \gain{4}
& 3 & 7 & \gain{4} \\

Disfluency/Correction
& 12.05 & 12
& 5 & 7 & \gain{2}
& 0 & 5 & \gain{5} \\

Indirect Speech
& 19.28 & 19
& 6 & 12 & \gain{6}
& 5 & 9 & \gain{4} \\

Referential Ambiguity
& 12.85 & 13
& 5 & 10 & \gain{5}
& 5 & 7 & \gain{2} \\

Under-specification
& 38.15 & 38
& 16 & 23 & \gain{7}
& 10 & 18 & \gain{8} \\

\midrule
\textbf{Overall}
& \textbf{100.00} & \textbf{100}
& \textbf{37} & \textbf{61}
& \cellcolor{gainbg}\textcolor{gain}{\bfseries +24}
& \textbf{23} & \textbf{46}
& \cellcolor{gainbg}\textcolor{gain}{\bfseries +23} \\

\bottomrule
\end{tabular}
}
\end{table}


\section{Experiments}
We conduct experiments to answer two questions:  (1) how well current GUI agents and VLMs execute naturally elicited older-adult instructions, and  (2) which linguistic and execution-level factors contribute to their failures. We first evaluate the performance of mainstream GUI agents and VLMs under ElderBench's online and offline protocols.
We then conduct controlled instruction normalization, trajectory-level failure analysis, and linguistic feature analysis to examine the sources of performance degradation.

\subsection{Evaluated Models and Settings}
To fully evaluate the performance of autonomous systems applied in elderly-oriented scenarios, we select a range of mainstream GUI agents and VLMs as evaluation subjects. The selected GUI agents cover AutoGLM \cite{liu2024autoglm}, Doubao-1.5-UI-TARS \cite{ui-tars-15-seed}, GUI-Owl-1.5 \cite{xu2026mobile} and UI-Venus-1.5 \cite{team2026ui}. Furthermore, we also evaluate VLMs for GUI navigation tasks, namely Qwen3-VL-Flash \cite{bai2025qwen3}, GLM-4.6V \cite{hong2025glm}, Qwen3-VL-Plus \cite{bai2025qwen3}, Gemini-3-Flash \cite{comanici2025gemini} and Kimi K2.5~\cite{team2026kimi}.

For the specialized frameworks (AutoGLM, Doubao-1.5-UI-TARS, GUI-Owl-1.5, and UI-Venus-1.5), we strictly adhere to their official open-source repositories and native implementation configurations to ensure an uncorrupted evaluation. Aligning with the dynamics execution of real-world smartphone operating systems and following the experimental setup by prior literature~\cite{xu2025androidlab}, we impose a strict execution budget for every task: the maximum trajectory length is bounded at $25$ operational steps, coupled with a hard wall-clock timeout threshold of $120$ seconds.
Full device configurations, app versions, account statuses and permission settings are in supplementary materials.

\subsection{Results and Analysis}
\label{sec:results_and_analysis}

Table~\ref{tab:mode_breakdown} reports the overall, online, and offline performance of the evaluated VLMs and GUI agents. The online subset contains 130 tasks across 15 applications, while the offline subset contains 119 tasks across five applications. Since the two subsets cover different tasks and applications, their results should be viewed as complementary rather than as a controlled comparison of evaluation protocols.

\textbf{Overall Performance.}
ElderBench remains challenging across both model categories. Among VLMs, Gemini-3-Flash achieves the highest overall SR of 49.80\%, followed by Qwen3-VL-Plus at 47.39\%. Among GUI agents, GUI-Owl-1.5 performs best with 45.38\% SR, followed by UI-Venus-1.5 at 41.37\%. No model exceeds 50\% overall SR, revealing substantial room for improvement on naturally elicited older-adult instructions.

\textbf{Online and Offline Evaluation.}
Models generally obtain higher SR online. Qwen3-VL-Plus and GUI-Owl-1.5 achieve the best online SRs among VLMs and GUI agents,
reaching 76.15\% and 68.46\%, respectively. In contrast, Gemini-3-Flash and UI-Venus-1.5 perform best offline, with SRs of
27.73\% and 31.09\%. The lower offline scores may partly result from the task-graph protocol, which requires each predicted action to match
a valid graph edge, whereas online agents can observe updated states and recover through alternative paths. However, the gap may also reflect
differences in task and application composition.

\textbf{Efficiency Trade-offs.}
No model simultaneously optimizes success, latency, and token cost. Qwen3-VL-Flash has the lowest overall AIL of 3.57 seconds but only
22.89\% SR and high token consumption. Gemini-3-Flash achieves the highest overall VLM SR with the lowest VLM token cost, while
UI-Venus-1.5 provides the best latency--accuracy balance among GUI agents. Offline MCM is consistently lower, whereas AIL varies across
models, suggesting that per-step latency is strongly model-dependent.

\subsection{Controlled Instruction Normalization}
\label{sec:instruction_normalization}
Overall performance gaps cannot isolate agent failures caused by senior-specific language from inherent GUI task difficulty. To disentangle the two factors, we run controlled instruction normalization with fixed tasks, device states, app environments and evaluation protocols.

We sample 100 stratified instructions following ElderBench’s five semantic categories: Clear and Explicit, Disfluency/Correction, Indirect Speech, Referential Ambiguity, Under-specification. All original instructions are rewritten into concise, action-focused variants while retaining task intent. The rewritten instructions were manually verified, with 90\% accepted directly and 10\% revised by the authors to ensure semantic consistency. 

We evaluate Qwen3-VL-Flash \cite{bai2025qwen3} and AutoGLM \cite{liu2024autoglm} on both original and rewritten instructions. As each rewritten instruction pairs with identical tasks and environments, performance differences quantifies the link between linguistic formulation and task completion.

\textbf{Effect of Instruction Normalization.}
As shown in Table~\ref{tab:instruction_normalization}, normalization increases successful tasks from 37 to 61 for AutoGLM and from 23 to 46 for Qwen3-VL-Flash. Both gains are significant under exact McNemar's test ($p<0.001$), with paired-bootstrap 95\% CIs of $[14,34]$ and $[12,34]$ percentage points. Gains appear across all categories, indicating that linguistic mismatch is a substantial failure source, alongside persistent planning, state-tracking, and environment-understanding challenges.

\subsection{Failure Analysis}
\label{sec:failure_analysis}

To identify the execution-stage bottlenecks behind task failures, we manually examined all 63 failed trajectories produced by AutoGLM on the 130-task online subset, where it achieved an SR of 51.54\%. AutoGLM was selected because it provides a complete closed-loop execution trace and represents a competitive GUI agent.

As shown in Table~\ref{tab:failure_categories}, failures are dominated by intent understanding and capability-boundary errors (57.1\%). These cases include interpreting indirect requests as knowledge questions, resolving vague references without sufficient evidence, and repeatedly searching for functions unavailable under the current application, permission, or action space.

Long-horizon planning and state-tracking errors account for another 33.3\%. Typical behaviors include omitting prerequisite steps, repeating completed operations, continuing after reaching the goal, and following the original plan after the observed interface has
diverged from the expected state. By comparison, device-environment awareness (6.4\%) and visual/action execution errors (3.2\%) occur less frequently. The former mainly involves assuming unavailable applications, accounts, or permissions, whereas the latter includes
mislocalized taps and incorrect structured actions.

Together, the first two categories account for 90.4\% of all analyzed failures, indicating that AutoGLM's primary bottlenecks in this subset lie in interpreting underspecified user goals and maintaining coherent multi-step execution, rather than in low-level visual grounding alone.

\begin{table}[t]
\centering
\small
\setlength{\tabcolsep}{3pt}
\caption{Primary causes of AutoGLM failures on the online subset
($n=63$).}
\label{tab:failure_categories}
\begin{tabularx}{\columnwidth}{Xrr}
\toprule
\textbf{Failure Category} & \textbf{Count} & \textbf{Percentage (\%)} \\
\midrule
Intent/Capability Boundary & 36 & 57.1 \\
Planning/State Tracking    & 21 & 33.3 \\
Environment Awareness      & 4  & 6.4 \\
Visual/Action Execution    & 2  & 3.2 \\
\midrule
Total & 63 & 100.0 \\
\bottomrule
\end{tabularx}
\end{table}

\subsection{Linguistic Feature Analysis}
\label{sec:in-depth analysis}
To explore execution bottlenecks of GUI agents in elderly-oriented scenarios, we perform fine-grained attribution analysis. We correlate elderly GUI instruction features with overall execution outcomes, separating single feature effects and cross feature influence. 
We provide additional implementation details in the supplementary material.

\subsubsection{Single Feature Association}
Our analysis reveals several linguistic characteristics associated with agent performance.


\textbf{(1) Challenges of Implicit and Incomplete Expressions.} Elliptical and Indirect Speech instructions are associated with lower success rates. These expressions often omit explicit operational predicates or describe desired states rather than concrete actions, requiring agents to infer missing action sequences. For example, agents are prone to failure on instructions such as ``I can't see it, make it bigger'' and ``The sound is too low, I cannot hear it,'' where they need to translate abstract user states into operations such as changing display or volume settings.
\textbf{(2) Scenario-dependent Performance Variation.} The impact of linguistic characteristics also varies across task scenarios. Information inquiry tasks generally achieve higher success rates due to relatively simple execution flows, whereas interaction-intensive scenarios such as System Settings and Tool Assistance present greater challenges.

\subsubsection{Cross Feature Interaction}
Agent failures are rarely associated with a single linguistic property; instead, they emerge from interactions among syntactic, semantic, and pragmatic factors. We identify three representative interaction patterns. 

\textbf{(1) Scenario-Specific Tolerance for Verbosity:} While excessive text length generally impairs execution, detailed descriptive expressions in “Information inquiry”-oriented tasks actually deliver effective restrictive keywords, improving the success rate of agent operations $(+0.70)$.
\textbf{(2) Complexity antagonism}: Long text paired with "System setting" tasks yields notable negative interaction $(-0.42)$. 
\textbf{(3) Linguistic attribute conflict}: "Canonical" and "Indirect Speech" show strong negative correlation $(-0.34)$, and their combination triggers conflicting effects.

\section{Design Insights}
\label{sec:insights}

Based on ElderBench results and linguistic analysis, we derive three design implications for elderly-oriented GUI agents.
\textit{(1) State-Aware Intent Understanding.}
Since older adults often express needs through implicit states rather than explicit actions, agents should infer user situations and translate them into executable operations.
\textit{(2) Proactive Clarification.}
The difficulty of Elliptical and Disfluency/Correction instructions suggests that agents should detect insufficient specifications and seek clarification before uncertain execution.
\textit{(3) Context-Aware Parsing.}
Because linguistic complexity has different effects across scenarios, agents should adapt instruction interpretation according to both language patterns and task contexts.

\section{Conclusion}


We introduce ElderBench, the first mobile GUI-agent benchmark built from naturally elicited older-adult instructions. Linguistic analysis reveals substantial differences from conventional benchmark language, while evaluations of mainstream GUI agents and VLMs expose persistent performance limitations. Controlled normalization, failure analysis, and feature-level attribution further identify linguistic mismatch and execution bottlenecks, informing the design of more adaptive and age-inclusive GUI agents.

\bibliography{custom}

\appendix

\section{Participant Demographics}
\label{app:participant}

We recruited 28 older adults to collect naturally elicited smartphone instructions for ElderBench. Participants were between 59 and 84 years old, including 8 males (28.57\%) and 20 females (71.43\%). To capture diverse smartphone experiences, participants varied substantially in usage duration and daily engagement.

Participants were recruited from multiple regions in China, including Shanghai, Anhui, Jiangsu, and Hebei. Although the current benchmark focuses on Chinese-speaking older adults due to practical collection constraints, the collected instructions cover heterogeneous daily usage scenarios rather than a single demographic profile.

Regarding smartphone experience, some participants had more than ten years of smartphone usage experience, whereas others had only 1--8 years of experience. Daily smartphone usage ranged from approximately 1--2 hours for light users to more than 5--6 hours for highly engaged users.

Before data collection, we explained the research objectives, data anonymization policy, and interview procedure. All participants provided verbal informed consent before participating.

\section{Data Collection Details}
\label{app:data_collection}

\subsection{Interview Protocol}

To obtain naturally occurring elderly GUI instructions, we conducted semi-structured interviews \cite{pierres2025exploring,liu2025envisioning} rather than providing predefined benchmark tasks. The goal was to minimize the influence of benchmark-style language and preserve older adults' spontaneous expressions.

The interview consisted of two stages.

\textbf{Stage 1: Existing smartphone usage.}
Participants were asked to describe smartphone activities they commonly performed in daily life. We introduced the concept of GUI agents and asked participants how such assistants could support their current mobile interactions.

\textbf{Stage 2: Desired assistance scenarios.}
Participants further described smartphone operations that they wished an agent could help accomplish, especially tasks that were difficult, confusing, or inconvenient.

During the entire elicitation process, no predefined task list, instruction templates, or linguistic examples were provided. This procedure avoids bias toward explicit benchmark-style commands and allows elderly-oriented linguistic patterns to naturally emerge.

\subsection{Instruction Filtering}

The interview process initially collected 251 candidate instructions. Three authors independently reviewed all collected instructions. Instructions that were impossible to execute because of missing goals or unrecoverable ambiguity were removed after discussion.

After filtering two invalid samples, ElderBench contains 249 executable GUI instructions.

\section{LLM-based Linguistic Annotation}
\label{app:annotation}

\subsection{Annotation Overview}

To characterize linguistic differences between ElderBench instructions and existing GUI benchmark instructions, we analyze instructions from three perspectives: syntax, semantics, and pragmatics. \textit{(Grounded in linguistic theory \cite{silverstein1972linguistic,kasirzadeh2023conversation})}.

We use MobileWorld \cite{kong2026mobileworld} instructions as the baseline instruction set. All baseline instructions are translated into the same language as ElderBench instructions before analysis to avoid language-related bias.

The annotation pipeline combines LLM-assisted extraction and human verification. DeepSeek-V4-Pro is used for preliminary linguistic analysis, including verb extraction, syntactic classification, semantic categorization, and under-specification detection.

The average agreement between LLM annotation results and manual verification reaches 93.10\%.

\subsection{Prompt Template}

The following prompt is used for extracting core operational verbs.

\begin{tcolorbox}[
colback=gray!5,
colframe=gray!50,
arc=2mm,
title=Prompt for Core Verb Extraction,
fonttitle=\bfseries]

\small
\ttfamily

You are a linguistics expert. Analyze the following GUI operation instruction and extract all core operational verbs.

Rules:

1. Only retain verbs representing actual interaction actions (e.g., open, send, search, connect).

2. Exclude auxiliary expressions or polite words.

3. Exclude application names and nouns.

4. If multiple verbs exist, separate them with commas. If no operational verb exists, output ``None''.

Instruction:
\{text\}

Output:

\end{tcolorbox}


\section{Extended Linguistic Analysis}
\label{app:linguistic_analysis}

This section provides additional statistics and implementation details for the linguistic analysis presented in the main paper.

\subsection{Syntactic Dimension}
\label{app:syntactic}

Following linguistic theory, we classify GUI instructions into four syntactic categories:







\begin{itemize}[
    leftmargin=*,
    itemsep=1pt,
    parsep=0pt,
    topsep=2pt
]

\item \textbf{Canonical:} Standard imperative instructions with explicit operational intentions.

\item \textbf{Narrative:} Declarative expressions describing user needs or situations rather than directly requesting an action.

\item \textbf{Elliptical:} Incomplete expressions that omit necessary execution information.

\item \textbf{Compound:} Instructions containing multiple intents or sequential/parallel operations.

\end{itemize}

Compared with baseline GUI instructions, ElderBench instructions show substantial structural differences.

Existing GUI benchmarks mainly contain explicit, action-oriented commands designed for evaluating complex workflows. In contrast, elderly-oriented instructions are shorter and more diverse in structure.

The detailed statistics are shown in Table~\ref{tab:syn_statistics}.






\begin{table}[t]
\centering
\small
\setlength{\tabcolsep}{3pt}

\caption{Syntactic statistics of ElderBench and baseline instructions.}
\label{tab:syn_statistics}

\begin{tabular}{lccc}
\toprule
\textbf{Dataset} &
\textbf{\#Instr.} &
\textbf{Avg. Length} &
\textbf{Avg. Verbs} \\
\midrule

ElderBench
& 249
& 15.89
& 1.26 \\

MobileWorld
& 201
& 88.09
& 2.82 \\

\bottomrule
\end{tabular}

\end{table}

\begin{figure}[t]
\centering
\includegraphics[width=\columnwidth]{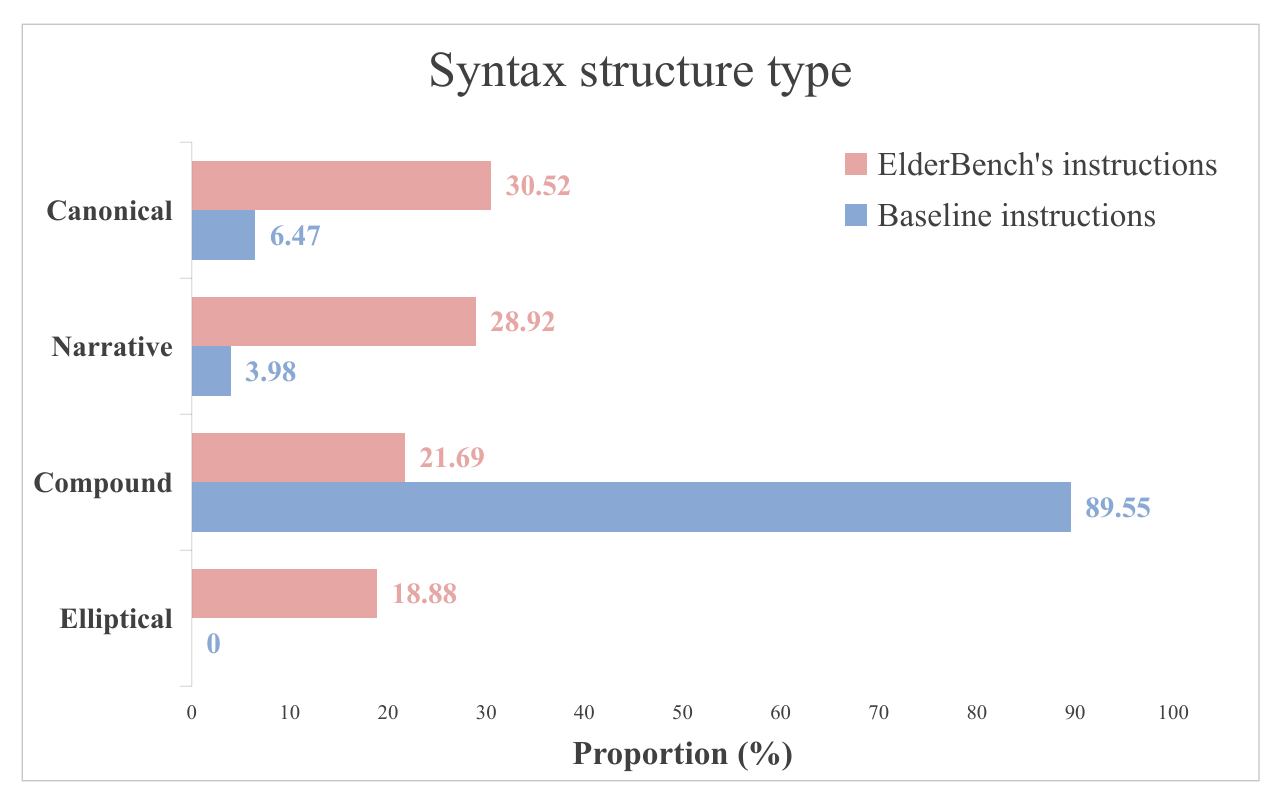}
\caption{
Distribution of syntactic structures. ElderBench instructions exhibit substantially higher proportions of Elliptical and Narrative expressions compared with baseline instructions.
}
\label{fig:syn_type}
\end{figure}

Figure~\ref{fig:syn_type} presents the distribution of four syntactic structures.

\begin{figure*}[t]
\centering
\includegraphics[width=\textwidth]{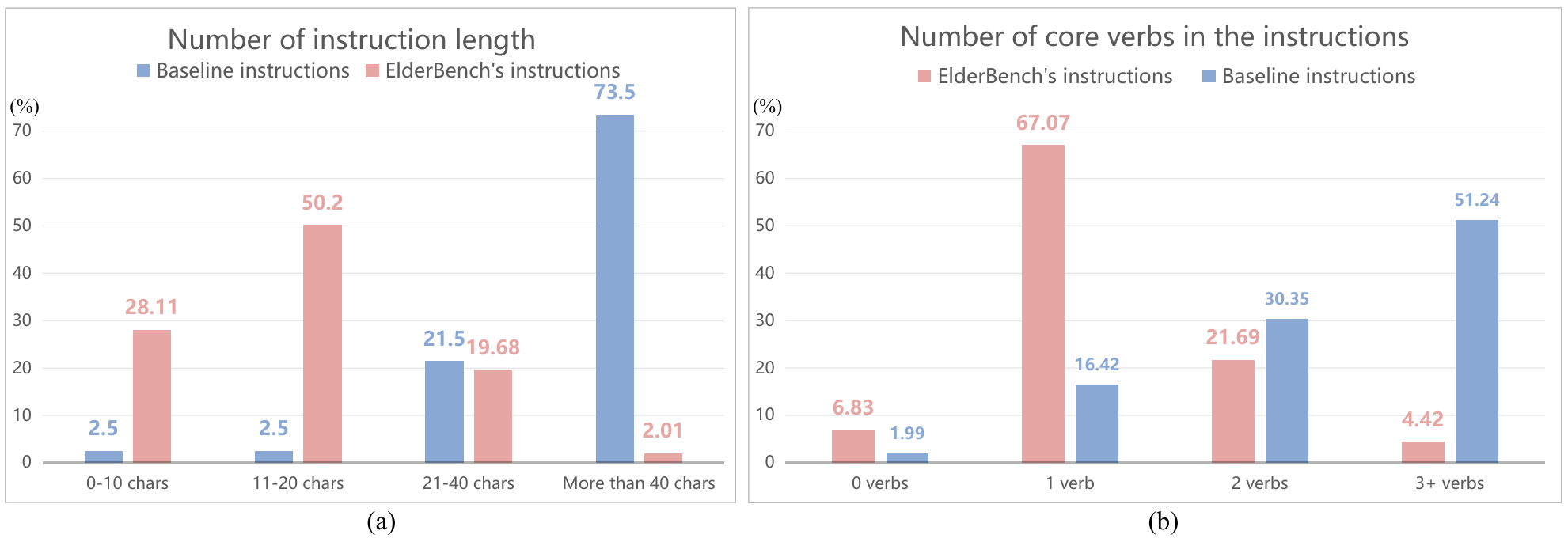}
\caption{
Comparison of instruction length and verb quantity distributions. Baseline instructions are generally longer and contain more explicit operational verbs, whereas ElderBench instructions are shorter and contain more implicit expressions.
}
\label{fig:syn_distribution}
\end{figure*}

Figure~\ref{fig:syn_distribution} further compares instruction length
and verb distributions.

\subsection{Semantic Dimension}
\label{app:semantic}

We perform thematic analysis \cite{clarke2017thematic} to identify semantic patterns beyond surface-level syntax.

Five semantic categories are considered:

\begin{itemize}[
    leftmargin=*,
    itemsep=1pt,
    parsep=0pt,
    topsep=2pt
]

\item \textbf{Clear and Explicit:}
The instruction directly specifies the intended action.

\item \textbf{Under-specification:}
The instruction expresses a clear goal but lacks required execution parameters.

\item \textbf{Disfluency/Correction:}
The instruction contains repetition, hesitation, or self-correction.

\item \textbf{Referential Ambiguity:}
The instruction relies on context-dependent references such as ``this'' or ``that''.

\item \textbf{Indirect Speech:}
The user describes a state or difficulty instead of explicitly stating the desired operation.

\end{itemize}

\begin{figure*}[t]
\centering
\includegraphics[width=\textwidth]{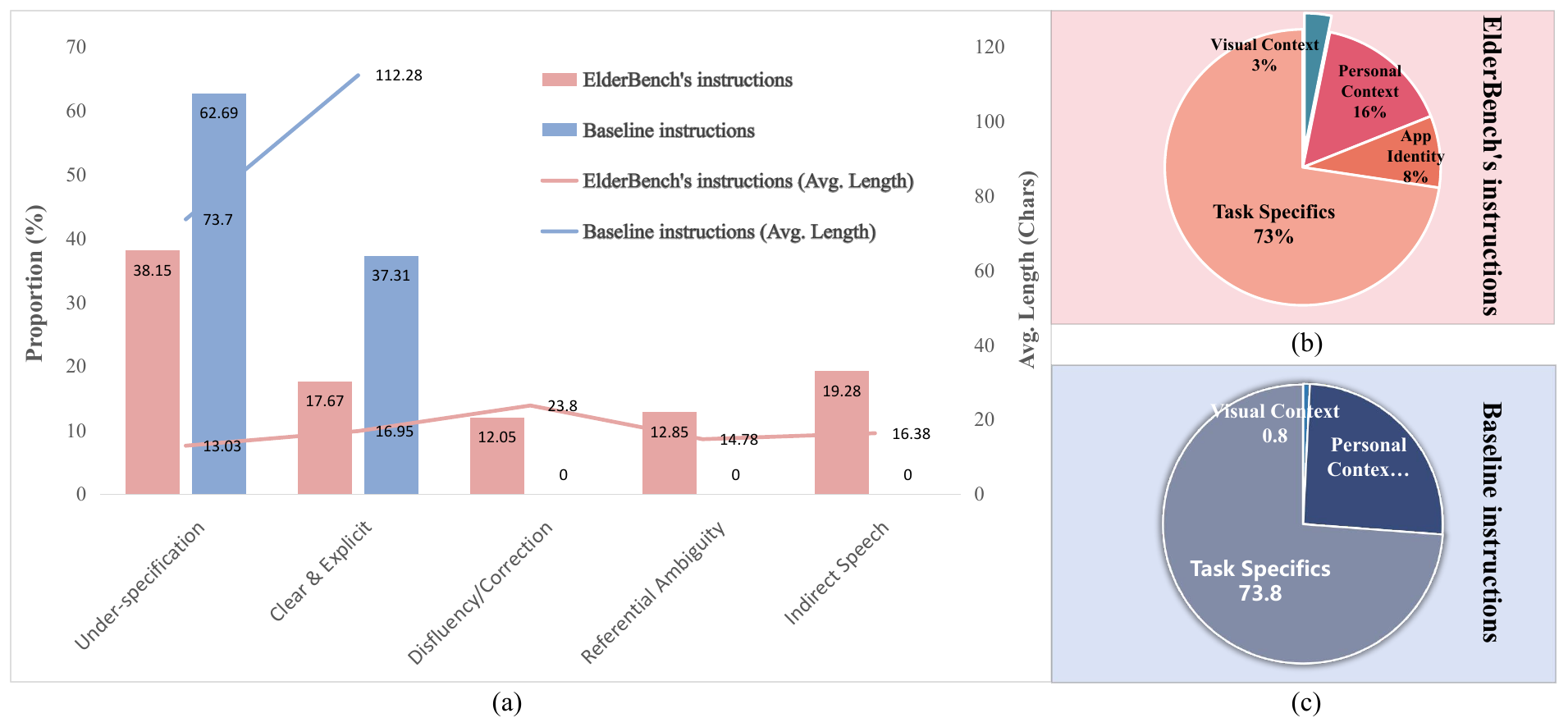}
\caption{
Semantic comparison between ElderBench and baseline instructions. Older adults frequently use indirect expressions and contextual references, while baseline instructions mainly contain explicit goals.
}
\label{fig:semantic_distribution}
\end{figure*}

The semantic distribution is shown in Figure~\ref{fig:semantic_distribution}.

\begin{figure*}[t]
\centering
\includegraphics[width=\textwidth]{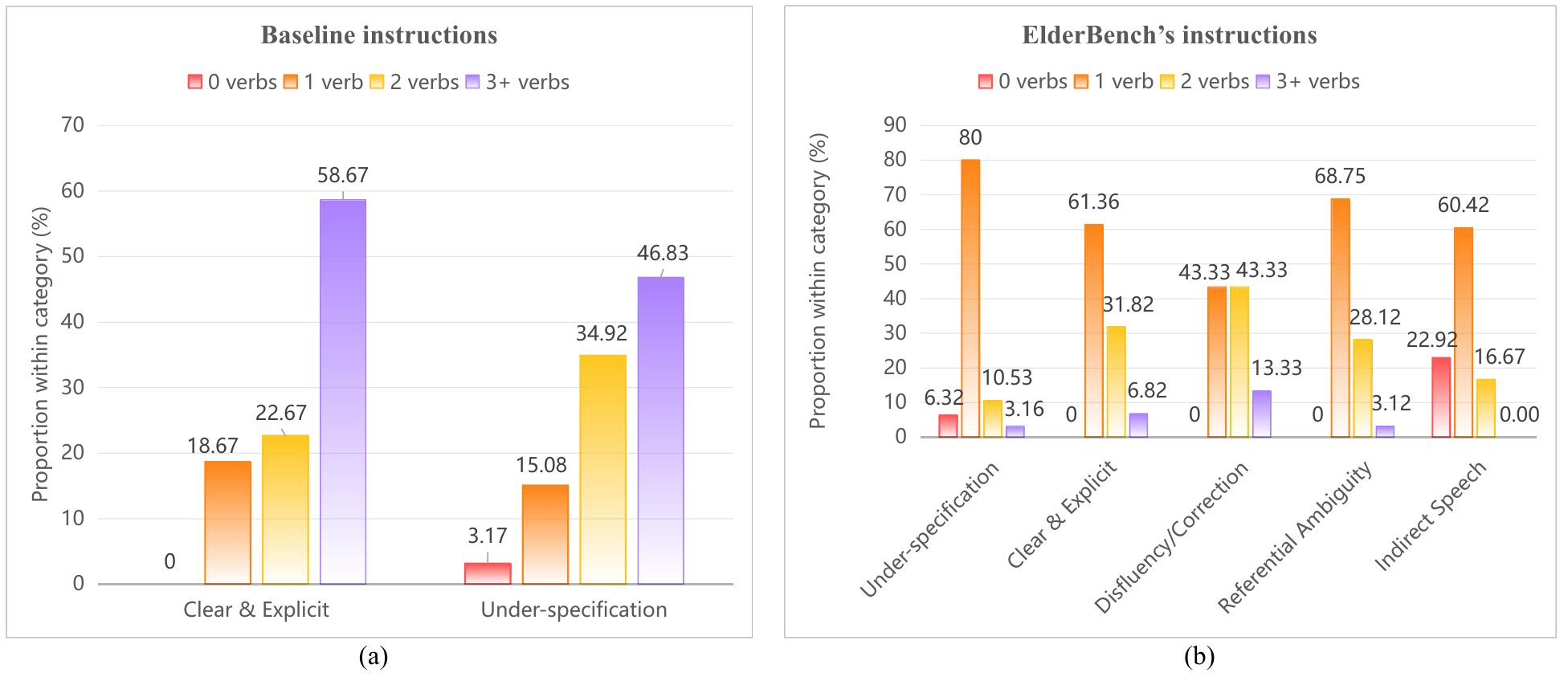}
\caption{
Verb distribution across semantic categories. Indirect Speech instructions contain a considerable proportion of zero-verb expressions.
}
\label{fig:semantic_verbs}
\end{figure*}

Additional analysis shows that indirect speech frequently lacks explicit operational verbs. Figure~\ref{fig:semantic_verbs} reports verb distribution across semantic categories.

\subsection{Pragmatic Dimension}
\label{app:pragmatic}

We analyze the application scenarios associated with elderly GUI instructions.

The collected instructions mainly concentrate on daily-life scenarios, including:

\begin{itemize}[
    leftmargin=*,
    itemsep=1pt,
    parsep=0pt,
    topsep=2pt
]

\item Life services: shopping, transportation, payment, and information retrieval.

\item Social communication: messaging, calling, and contact management.

\item System settings:
font adjustment, volume control, storage management, and accessibility configuration.

\end{itemize}

\begin{figure}[t]
\centering
\includegraphics[width=\columnwidth]{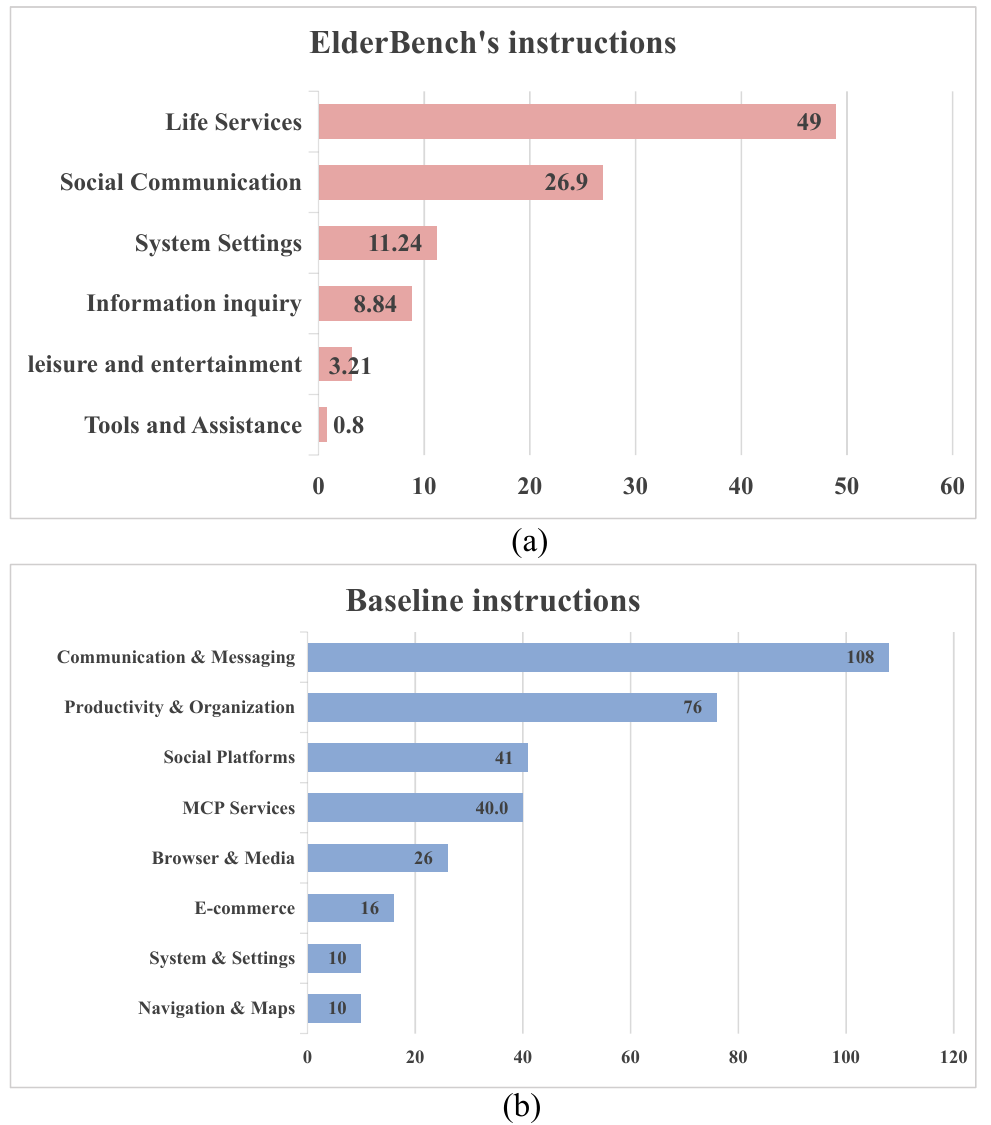}
\caption{
Distribution of pragmatic scenarios in ElderBench.
}
\label{fig:pragmatic_distribution}
\end{figure}

The scenario distribution is presented in Figure~\ref{fig:pragmatic_distribution}.

\section{Extended Feature Attribution Analysis}
\label{app:feature_analysis}

To further investigate how linguistic characteristics influence GUI agent performance, we conduct feature attribution analysis using AutoGLM as the representative GUI agent.

\subsection{Single Feature Influence}

We calculate Pearson correlation coefficients between individual instruction features and task success.

\begin{figure}[t]
\centering
\includegraphics[width=\columnwidth]{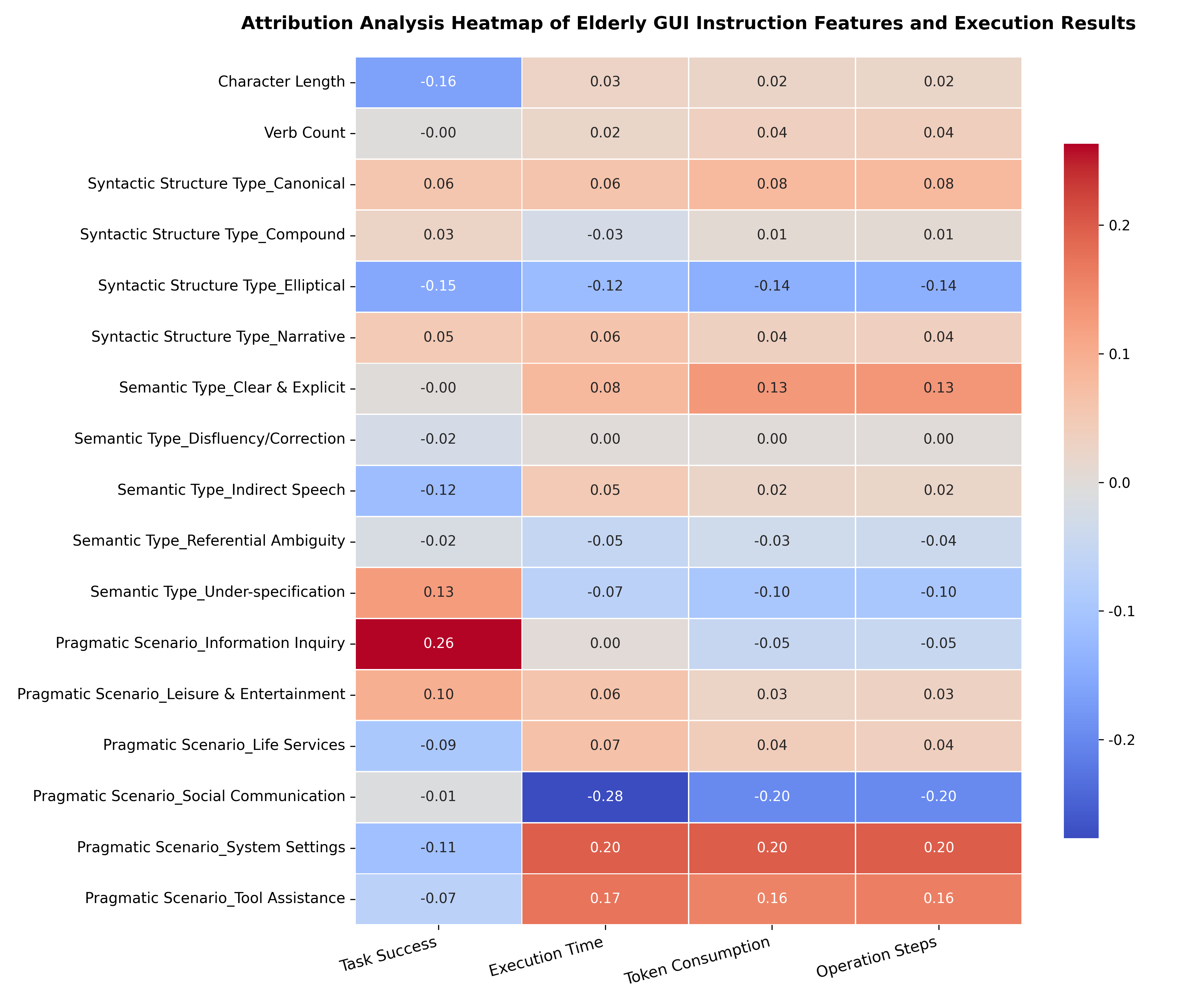}
\caption{
Single-feature correlation between linguistic attributes and task success.
}
\label{fig:single_heatmap}
\end{figure}

The resulting correlation matrix is shown in Figure~\ref{fig:single_heatmap}.

The analysis reveals several important trends:

\textbf{Instruction length.}
Longer instructions generally correlate negatively with task success ($-0.16$), suggesting that excessive descriptions may increase planning difficulty.

\textbf{Syntactic patterns.}
Elliptical structures produce the strongest negative correlation ($-0.15$), followed by Indirect Speech ($-0.12$).

\textbf{Scenario differences.}
Information inquiry tasks show positive correlation ($+0.26$), whereas System Settings ($-0.11$) and Tool Assistance ($-0.07$) introduce additional challenges.

\subsection{Cross Feature Influence}

To capture nonlinear feature interactions, we employ a Factorization Machine (FM) model \cite{rendle2010factorization}.

Categorical features, including semantic categories, syntactic structures, and scenarios, are encoded using one-hot representation. Continuous variables, including instruction length and verb number, are standardized.

The latent dimension is set to $k=5$. We train the model using Adam optimization with L2 regularization ($weight\ decay=0.001$).

The learned interaction matrix is calculated as:

\[
M=VV^{T}
\]

where each element represents the second-order interaction between two features.

Three representative interactions are:

\textbf{Scenario-specific tolerance for verbosity (+0.70).}

Although instruction length alone negatively affects performance, longer descriptions provide useful constraints in information inquiry tasks.

\textbf{Complexity clash (-0.42).}

Long descriptions combined with System Settings tasks introduce additional planning difficulty.

\textbf{Linguistic contradiction (-0.34).}

Canonical syntactic forms combined with Indirect Speech semantics create conflicting interpretation signals.


\section{Benchmark Implementation Details}
\label{app:benchmark_details}

\subsection{Overall Benchmark Statistics}




ElderBench contains 249 executable smartphone GUI tasks collected from older adults across 20 Android applications. The benchmark adopts complementary online and offline evaluation protocols. The online subset contains 130 tasks across 15 applications, as listed in Table~\ref{tab:online_apps}, while the offline subset contains 119 tasks across five applications: Alipay, WeChat, Gaode Map, Taobao, and Ctrip.


\begin{table}[t]
\centering
\small
\caption{Applications included in the online evaluation subset.}

\begin{tabular}{lll}
\toprule
\multicolumn{3}{c}{\textbf{Online Applications}}\\
\midrule
System Settings & Toutiao & Dianping \\
Weather & App Store & Douyin \\
Tencent News & Browser & Phone \\
Gallery & SMS & Meituan \\
Doubao & Alarm Clock & QQ Music \\
\bottomrule
\end{tabular}

\label{tab:online_apps}
\end{table}



The division is determined at the task level based on three criteria:

(1) reproducibility of execution environment;

(2) privacy and account-related constraints;

(3) stability of external application states.

Applications involving sensitive personal information, payment operations, or highly dynamic content are evaluated offline through human-demonstrated task graphs. Applications with relatively stable interaction flows are evaluated online through live interaction.

\section{Online Evaluation Environment}
\label{app:online_environment}

\subsection{Device Configuration}

The online evaluation is conducted on an Android emulator configured as shown in Table~\ref{tab:online_device}, with additional support for deployment on physical devices.

\subsection{Application Configuration}

Before each evaluation:

\begin{itemize}[
    leftmargin=*,
    itemsep=1pt,
    parsep=0pt,
    topsep=2pt
]

\item Search history was cleared;

\item Location permission was enabled when required;

\item Application states were reset;

\item No personalized historical content was retained.

\end{itemize}

Representative online application versions are listed in Table~\ref{tab:online_apps_versions}.

\begin{table}[t]
\centering
\small

\caption{
Representative online application versions.
}
\begin{tabular}{ll}
\toprule
Application & Version\\
\midrule

Meituan & 12.57.203\\
Gaode Map & 16.17.0.2003\\
Doubao & 13.4.0\\
QQ Music & 20.4.1.8\\
Douyin & 38.8.0\\
Toutiao & 16.8.0\\
Dianping & 11.63.3\\

\bottomrule
\end{tabular}

\label{tab:online_apps_versions}

\end{table}

\section{Offline Task Graph Construction}
\label{app:task_graph}

\subsection{Human Demonstration Collection}

The offline evaluation follows a human-demonstration-based trajectory construction process.

For each offline task, human annotators execute the intended operation on physical devices while recording interaction trajectories.

At each timestep $t$, we collect:
\[
(O_t,V_t,A_t)
\]
where:

\begin{itemize}[
    leftmargin=*,
    itemsep=1pt,
    parsep=0pt,
    topsep=2pt
]

\item $O_t$ denotes the screenshot observation;

\item $V_t$ denotes the UI hierarchy extracted through UIAutomator;

\item $A_t$ denotes the executed action, including action type,
coordinates, and text input.

\end{itemize}

The initial trajectory is designed according to official application functions to ensure that the operation path is valid and executable.


\begin{table}[t]
\centering
\small
\caption{
Online evaluation environment configuration.
}

\begin{tabular}{ll}
\toprule
Component & Configuration\\
\midrule

Device model & Google Pixel 6a\\
Operating System & Android 17\\
CPU & 4 cores\\
Memory & 6GB RAM\\
Storage & 16GB\\
Resolution & 1080 $\times$ 2400\\

\bottomrule
\end{tabular}

\label{tab:online_device}

\end{table}

\subsection{Multi-path Task Graph Generation}

To avoid evaluating agents against a single ``golden trajectory'', we collect multiple valid demonstrations for each task.

The collected trajectories are merged into a task graph through the following procedure:

\begin{enumerate}[
    leftmargin=*,
    itemsep=1pt,
    parsep=0pt,
    topsep=2pt
]

\item Extract UI states from all demonstration trajectories.

\item Match equivalent states using page signatures, anchor texts,
package names, and UI component information.

\item Merge equivalent states into graph nodes.

\item Convert human actions between states into directed edges.

\item Mark successful terminal states as DONE nodes.

\end{enumerate}

The final graph represents multiple valid execution paths.

During evaluation, an agent action is considered valid only when it matches an available outgoing edge from the current graph node.

For click actions, a spatial tolerance of 50 pixels is allowed. Text inputs are evaluated using fuzzy matching.

A task is successful only when:
$
Agent\_Finish \cap Graph\_DONE
$
is satisfied.

\section{Evaluation Setting Details}
\label{app:evaluation_setting}

\subsection{Single-turn Evaluation}

ElderBench focuses on single-turn GUI task execution.

This setting is motivated by three observations:

First, as the first benchmark specifically targeting elderly-oriented GUI interaction, ElderBench establishes a fundamental evaluation scenario based on naturally collected instructions.
Second, our interviews show that older adults frequently express digital needs as standalone requests. Third, most current mobile GUI agents operate primarily on single-turn user instructions; therefore, single-turn evaluation best reflects the present state of the technology.



\section{Controlled Instruction Normalization}
\label{app:normalization}

To examine whether elderly-specific linguistic expressions contribute to agent failures, we conduct a controlled instruction normalization study.

We sample 100 instructions according to the semantic distribution of ElderBench. Each original instruction is rewritten into a concise, explicit, action-oriented form while preserving the original task goal.

The rewritten instructions are manually verified to ensure semantic equivalence.

\begin{table}[t]
\centering
\small

\caption{
Semantic distribution of the normalization subset.
}

\begin{tabular}{lcc}
\toprule
Category &
ElderBench(\%) &
Sample Size\\
\midrule

Clear and Explicit
&17.67&18\\

Disfluency/Correction
&12.05&12\\

Indirect Speech
&19.28&19\\

Referential Ambiguity
&12.85&13\\

Under-specification
&38.15&38\\

\midrule
Total
&100&100\\

\bottomrule
\end{tabular}

\label{tab:norm_distribution}

\end{table}

We evaluate Qwen3-VL-Flash \cite{bai2025qwen3} and AutoGLM (Phone-9B) \cite{liu2024autoglm} on both original and normalized instructions.

Because each pair corresponds to the same task and environment, the performance difference estimates the effect of linguistic formulation rather than task difficulty.

\section{Failure Analysis}
\label{app:failure_analysis}

To understand why GUI agents fail on elderly-oriented instructions, we manually inspect failed trajectories of AutoGLM on the online subset.

Among 63 failed tasks, we categorize failures into four major types.

\begin{figure}[t]
\centering
\includegraphics[width=\columnwidth]{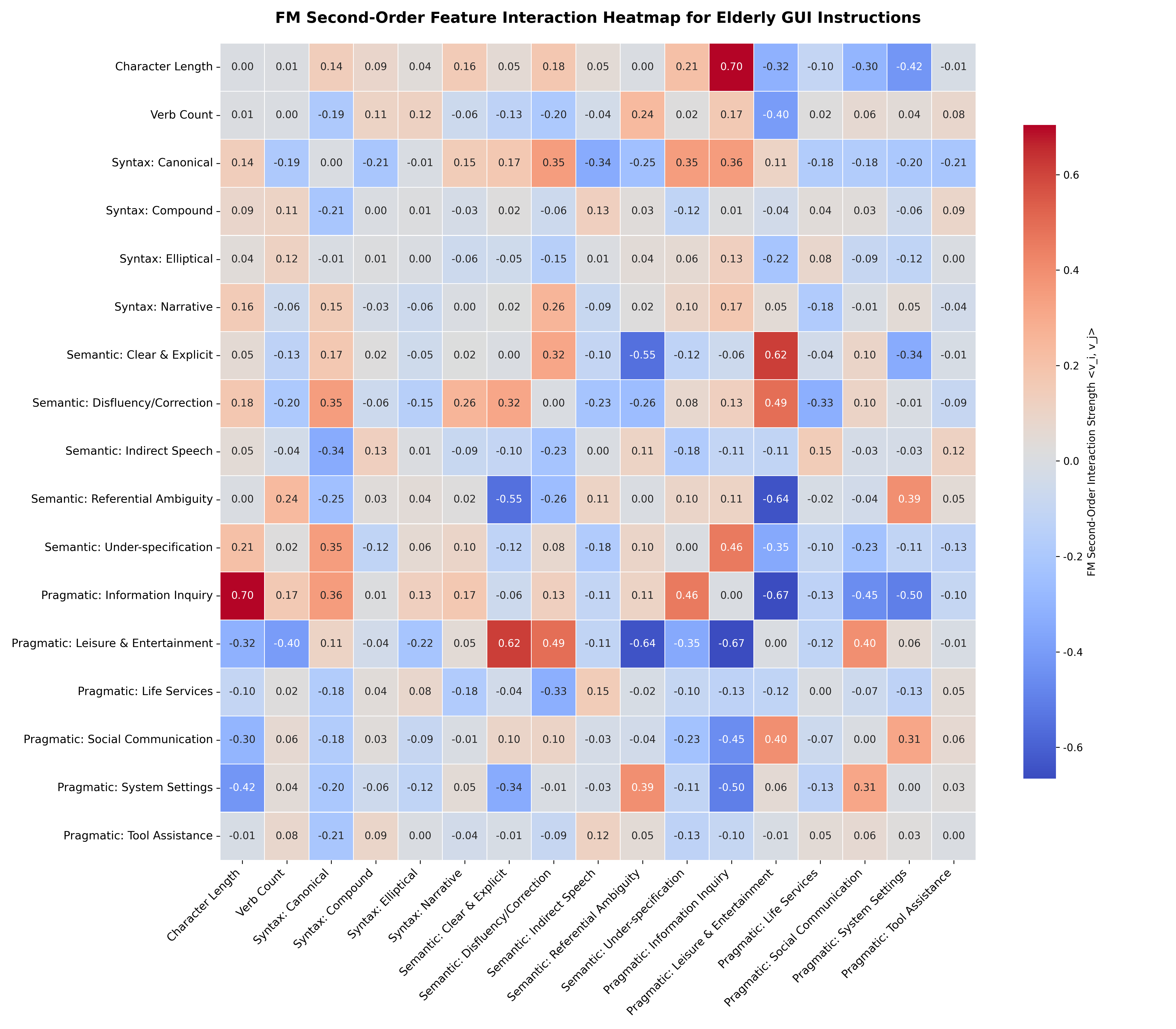}
\caption{
Second-order feature interactions learned by the Factorization Machine.
}
\label{fig:fm_heatmap}
\end{figure}

\begin{table}[t]
\centering
\small
\setlength{\tabcolsep}{3pt}
\caption{Failure categories of AutoGLM on online evaluation ($n=63$).}
\label{tab:failure_categories}
\begin{tabularx}{\columnwidth}{Xrr}
\toprule
\textbf{Failure Category} & \textbf{Count} & \textbf{Percentage (\%)} \\
\midrule
Intent/Capability Boundary & 36 & 57.1 \\
Planning/State Tracking    & 21 & 33.3 \\
Environment Awareness      & 4  & 6.4 \\
Visual/Action Execution    & 2  & 3.2 \\
\midrule
Total & 63 & 100.0 \\
\bottomrule
\end{tabularx}
\end{table}

\subsection{Intent Understanding and Capability Boundary}

This category represents failures caused by incorrect interpretation of elderly expressions or inability to determine whether a requested operation is feasible.

Typical examples include:

\begin{itemize}[
    leftmargin=*,
    itemsep=1pt,
    parsep=0pt,
    topsep=2pt
]

\item interpreting ``the sound is too small'' as a question rather than a volume adjustment request;

\item failing to resolve references such as ``turn this off'';

\item attempting unavailable operations requiring system-level permissions.

\end{itemize}

\subsection{Planning and State Tracking}

These failures occur when agents cannot maintain task progress during multi-step interaction.

Common patterns include:

\begin{itemize}[
    leftmargin=*,
    itemsep=1pt,
    parsep=0pt,
    topsep=2pt
]

\item repeating completed operations;

\item missing intermediate steps;

\item continuing execution after reaching the goal.

\end{itemize}

\subsection{Environment Awareness}

Some agents rely excessively on pretrained application knowledge and fail to verify the actual device environment.

Examples include selecting unavailable applications or ignoring required permissions.

\subsection{Visual Localization and Action Execution}

These failures occur when the agent correctly understands the goal but cannot accurately locate UI elements or generate correct interaction coordinates.

\section{Additional Implementation Details}

The maximum execution trajectory length is limited to 25 operational steps. Each task has a 120-second timeout threshold.

All evaluated GUI agents interact with Android environments through ADB commands. The supported action space includes:

\begin{itemize}[
    leftmargin=*,
    itemsep=1pt,
    parsep=0pt,
    topsep=2pt
]

\item Tap

\item Long Press

\item Swipe

\item Type

\item Home / Back

\end{itemize}

The coordinate system is normalized to $[0,1000]$ and automatically converted into device-specific pixel coordinates.



\end{document}